\documentclass[runningheads]{llncs}

\usepackage{eccv}

\usepackage{eccvabbrv}

\usepackage{graphicx}
\usepackage{booktabs}

\usepackage[accsupp]{axessibility}  

\usepackage{hyperref}

\usepackage{orcidlink}

\usepackage{siunitx}
\usepackage{texnames}
\usepackage{bm,bbm}
\usepackage{orcidlink}

\usepackage{booktabs}
\usepackage{graphicx}

\usepackage{tikz}
\usetikzlibrary{positioning}
\usetikzlibrary{calc}

\usepackage[percent]{overpic}

\usepackage{multirow}
\usepackage{adjustbox}
\usepackage[table]{xcolor}

\DeclareMathAlphabet{\mathcal}{OMS}{cmsy}{m}{n}

\begin{document}

\title{Robust RPC Bundle Adjustment for Multi-Date Satellite Imagery with Season-Invariant Correspondences} 

\titlerunning{Robust RPC Bundle Adjustment for Multi-Date Satellite Imagery}

\author{Roger Marí\inst{1}\and
Elías Masquil\inst{2}\and\\
Xavier Bou\inst{3}\and
Thibaud Ehret\inst{3}\and
Gabriele Facciolo\inst{4,5}
}

\authorrunning{R.~Marí et al.}

\institute{\textsuperscript{1 }Eurecat, Multimedia Technologies, Barcelona, Spain \\
\textsuperscript{2 }IIE, Facultad de Ingeniería, Universidad de la República, Uruguay \\
\textsuperscript{3 }AMIAD, France \qquad \textsuperscript{4} Institut Universitaire de France \\
\textsuperscript{5 }Université Paris-Saclay, CNRS, ENS Paris-Saclay, Centre Borelli, France \\
\email{roger.mari@eurecat.org}}

\maketitle

\begin{abstract}
Accurate refinement of Rational Polynomial Camera (RPC) models is essential for high-quality satellite image geolocation. In ground control point (GCP)-free multi-view pipelines, this refinement is commonly performed through bundle adjustment from automatically extracted image correspondences. However, conventional RPC bundle adjustment pipelines rely on handcrafted feature matching, which becomes unreliable in multi-date collections affected by seasonal, illumination, and land-cover changes. We propose an appearance-aware RPC refinement pipeline that combines learned local feature matching for season-invariant correspondences with global image descriptors for selecting visually compatible image pairs. This reduces redundant and error-prone matching while preserving the connectivity of the matching graph.
Experiments on seasonally diverse WorldView-3 images show that our pipeline improves GCP-free relative RPC refinement over open-source baselines, achieving lower geometric consistency errors while substantially reducing matching time on collections with 39-42 views.
By making RPC refinement more robust to diachronic appearance variation, our approach enables more effective use of multi-date satellite imagery.

\keywords{RPC models \and Multi-date satellite imagery \and Bundle adjustment \and Tie-point extraction \and Season-invariant correspondences}

\end{abstract}

\tikzset{
  figlabel/.style={font=\fontfamily{ptm}\selectfont\footnotesize}
}

\newcommand{\smalltilde}{\raisebox{0.15ex}{\tiny$\sim$}}
\begin{figure}[t]
\centering
  \includegraphics[
    draft=false,
    width=\linewidth
  ]{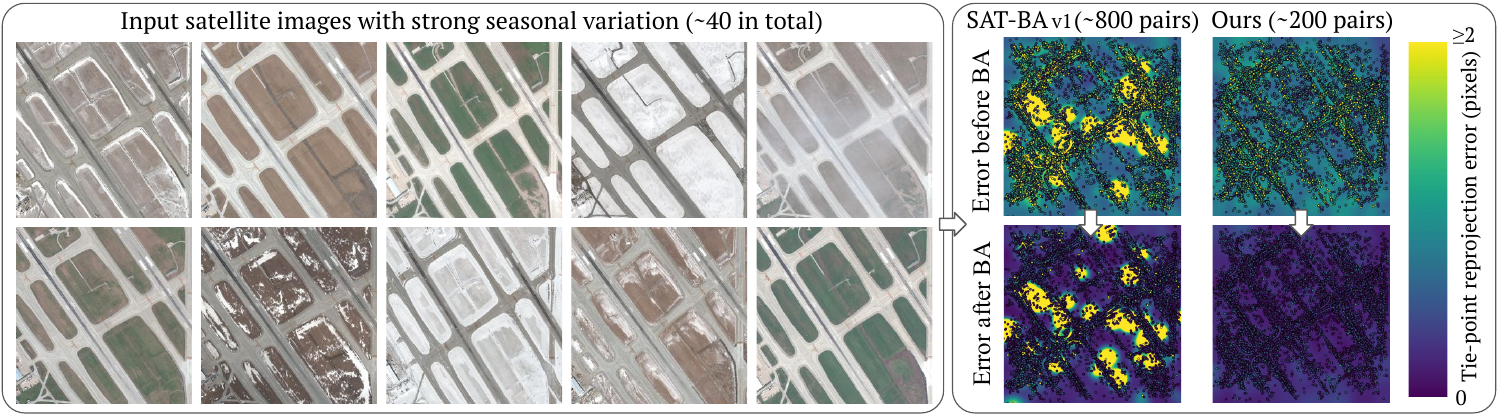}
\caption{The proposed RPC bundle adjustment (BA) pipeline handles appearance changes in multi-date satellite imagery using season-invariant correspondences and avoiding error-prone pairwise matching. Compared with existing baselines such as \mbox{SAT-BA{\footnotesize v1}}~\cite{mari2021generic}, it achieves lower reprojection errors, indicating improved geometric accuracy, while using fewer matched image pairs. Area of interest: \textit{OMA~059}, DFC2019 \textit{Track3}~\cite{lesaux20192019}. Left: input images. Right: tie-point reprojection-error heat maps before and after BA, mapped onto the geographic footprint of the area.}
\label{fig:teaser}
\end{figure}

\section{Introduction}
\label{sec:intro}

The refinement of Rational Polynomial Camera (RPC) models is a critical step for improving the geometric accuracy of high-resolution satellite imagery~\cite{grodecki2001ikonos, grodecki2003block}. RPCs define the mapping from 3D object space to 2D image coordinates, and are central to geolocating satellite images and downstream tasks such as image orthorectification, registration, change detection, and 3D reconstruction~\cite{hu2004understanding}. Bundle adjustment methods are commonly adopted for RPC refinement in multi-view satellite image pipelines~\cite{beyer2018ames, mari2021generic, leotta2019urban, aati2020optimization, rupnik2016refined}, where camera correction parameters are optimized by minimizing the reprojection errors of image correspondences, commonly referred to as tie points.

Existing satellite RPC bundle adjustment pipelines rely on handcrafted feature matching for automatic tie-point extraction~\cite{beyer2018ames, mari2021generic, aati2020optimization, ling2021unified}, implicitly assuming consistent, synchronic image appearance across views. This assumption breaks down in diachronic multi-date collections, where seasonal, illumination, and land-cover changes introduce strong appearance variability, degrading the performance of conventional pairwise matching strategies~\cite{masquil2026diachronic}. As a result, noisy or incorrect tie points may contaminate the bundle adjustment optimization, leading to suboptimal or unstable RPC corrections.

In this work, we propose a robust and scalable bundle adjustment pipeline to extend RPC refinement to diachronic multi-date satellite image collections.
This setting is increasingly relevant for large Earth observation archives, where images of the same area are acquired across seasons, years, and changing surface conditions. Our approach reduces redundant, error-prone pairwise matching and improves tie-point reliability (Fig.~\ref{fig:teaser}).
The main contributions of this work are:

\begin{itemize}
\item An appearance-aware tie-point extraction strategy for bundle adjustment that replaces handcrafted feature matching with learned detector-based matching, improving robustness under strong appearance changes.
\item A scalable image-pair selection strategy based on global visual similarity, designed to prioritize visually compatible pairs and filter out pairs with strong seasonal discrepancies, while preserving the matching graph connectivity.
\end{itemize}
We validate the proposed pipeline on seasonally diverse WorldView-3 image collections through feature matching and image descriptor benchmarks, ablation studies, and comparisons with open-source RPC correction pipelines~\cite{beyer2018ames, mari2021generic}. Experimental results show that our method improves RPC refinement accuracy while reducing runtime, providing a practical pathway for exploiting large-scale multi-date satellite image archives with reliable relative geometric consistency. Project page:
\url{https://centreborelli.github.io/sat-bundleadjust/v2}.
\section{Related Work}
\label{sec:literature}

\subsection{RPC Functions in a Nutshell}

The Rational Polynomial Camera (RPC) model is a generic sensor model widely used to describe the geometry of remote-sensing images independently of the physical sensor design. It defines a projection function $\mathcal{P}: \mathbb{R}^3 \rightarrow \mathbb{R}^2$ from object space to image coordinates, and a localization function $\mathcal{L}: \mathbb{R}^2 \times \mathbb{R} \rightarrow \mathbb{R}^3$ that maps an image point and height hypothesis to object space. Both functions are expressed as ratios of cubic polynomials~\cite{grodecki2003block,tao2001sensor}, e.g.,
\begin{equation}
    \mathbf{x} = \mathcal{P}(\mathbf{X})=\left( \frac{a(\mathbf{X})}{b(\mathbf{X})}, \frac{e(\mathbf{X})}{f(\mathbf{X})} \right),
    \label{eq:rpc_projection}
\end{equation}
where $\mathbf{X}$ and $\mathbf{x}$ are normalized object- and image-space coordinates, respectively, and $a, b, e, f$ are 20-coefficient cubic polynomials, yielding 80 coefficients for $\mathcal{P}$ and 80 for $\mathcal{L}$, plus the corresponding normalization offsets and scales.

RPC models enable surface reconstruction by triangulating 2D correspondences across images into 3D scene points. However, vendor-provided RPCs often contain systematic biases that limit high-accuracy geopositioning. The dominant source of error is typically uncertainty in the satellite attitude angles, which determine the sensor orientation~\cite{grodecki2003block,fraser2005bias,tong2010bias}. As a result, a 3D point may reproject several pixels away from its true image location, leading to geometric inconsistencies between overlapping views of the same geographic area.

\subsection{RPC Correction Methods} 

Absolute RPC correction requires external georeferenced data, such as ground control points (GCPs), i.e., image observations with known object-space coordinates~\cite{tao2001sensor,grodecki2003block,xiong2009generic,tong2010bias}. Digital elevation models (DEMs)~\cite{cao2019bundle} or reference images~\cite{dangelo2008towards} can also be used to anchor RPCs in a global frame. In contrast, relative RPC correction relies primarily on tie points detected between overlapping images~\cite{wohlfeil2012fully,gong2017relative,mari2021generic,ban2024rational}, making it well suited to automatic inter-image alignment and multi-view reconstruction without requiring external data~\cite{zhang2019leveraging,mari2019bundle}. 

{\bf Direct correction methods} explicitly adjust the RPC polynomial coefficients~\eqref{eq:rpc_projection}, providing maximum flexibility but introducing a high-dimensional and redundant parameterization. As a result, direct refinement is susceptible to instability and overfitting when only limited GCPs or tie points are available~\cite{xiong2009generic,tong2010bias}. Consequently, direct RPC correction is less common than indirect correction in multi-view satellite pipelines.

{\bf Indirect correction methods} estimate auxiliary correction functions that are composed with the original RPC. These methods use a small set of additional parameters to capture geometric errors without modifying the RPC coefficients.

A first family of indirect methods operates in the image domain. In this setting, the corrected projection is obtained by applying a 2D transformation $\mathcal{T}: \mathbb{R}^2 \rightarrow \mathbb{R}^2$ after the original RPC projection. The simplest model is a 2D translation, also termed bias or correction offset, which compensates the dominant effect of attitude inaccuracies over moderate-size satellite image blocks~\cite{grodecki2003block,fraser2005bias,dangelo2012dsm}. More general affine or polynomial functions in image space can also compensate additional residual inconsistencies~\cite{rupnik2016refined}.

A second family of indirect methods operates in object space, applying the correction before the original RPC projection by composing $\mathcal{P}$ with a 3D transformation $\mathcal{T}: \mathbb{R}^3 \rightarrow \mathbb{R}^3$. Since attitude inaccuracies mainly induce small sensor-orientation errors, these methods typically refine each RPC by estimating a corrective rotation in object space~\cite{beyer2018ames,rupnik2016refined,mari2019bundle,mari2021generic}. Compared with image-space offsets, object-space corrections are more physically constrained and can better preserve geometric consistency beyond a limited region or height range~\cite{hanley2004sensor,fraser2005bias}.

\subsection{Bundle Adjustment}

Bundle adjustment is the predominant strategy for indirect RPC correction. It jointly minimizes the reprojection errors of tie-point correspondences with respect to the correction parameters and, in GCP-free relative settings, the 3D tie-point coordinates~\cite{triggs1999bundle, grodecki2003block}. Although bundle adjustment is a common pre-processing step in multi-view satellite pipelines~\cite{beyer2018ames, leotta2019urban, zhang2019leveraging}, few open-source tools expose both the correction parameters and the corrected camera models in standard RPC format, limiting systematic evaluation and compatibility with downstream geospatial software.
This constraint motivates our focus on two open-source pipelines that provide the required outputs: the \textit{Ames Stereo Pipeline} (ASP)~\cite{beyer2018ames} and \textit{sat-bundleadjust}~v1 (SAT-BA{\footnotesize v1})~\cite{mari2021generic}.

Both ASP and SAT-BA{\footnotesize v1} model RPC refinement as a 3D transformation $\mathcal{T}$ applied in object space before the original RPC projection $\mathcal{P}$. For ASP, ECEF object-space coordinates $\mathbf{X}$ are transformed as
\begin{equation}
    \mathbf{X}' = \mathcal{T}_{\mathrm{ASP}}(\mathbf{X}) = R(\mathbf{X}-\mathbf{C}) + \mathbf{C} + \mathbf{T},
    \label{eq:asp_formula}
\end{equation}
where $R$ is a rotation matrix, $\mathbf{T}$ a translation vector, and $\mathbf{C}$ a fixed approximate camera center used as the rotation center. 
SAT-BA{\footnotesize v1} uses a rotation-only correction model, since translations were found to be largely redundant with rotations in the satellite imaging geometry, owing to the large sensor-to-scene distance and the small viewing-angle corrections:
\begin{equation}
    \mathbf{X}' = \mathcal{T}_{\mathrm{SAT\text{-}BA_{v1}}}(\mathbf{X}) = R(\mathbf{X}-\mathbf{C}) + \mathbf{C}.
    \label{eq:sat-ba_formula}
\end{equation}
Thus, both methods apply a corrective rotation $R$, while ASP additionally estimates a translation term in~\eqref{eq:asp_formula}. After bundle adjustment, the corrected mapping is converted back to RPC form by fitting a new model from dense correspondences obtained by sampling 3D points and projecting them through $\mathcal{P}\circ\mathcal{T}$~\cite{akiki2021robust}.

Other open-source pipelines replace RPC projection with locally valid approximations. VisSat approximates RPC cameras as local perspective cameras and refines the projection matrices through bundle adjustment~\cite{zhang2019leveraging}, enabling compatibility with pinhole-based SfM and MVS pipelines such as COLMAP~\cite{schoenberger2016sfm, schonberger2016pixelwise}. However, the corrected cameras are no longer represented as RPC models.

\subsection{Assessment of Corrected RPC Models}

The standard protocol for assessing RPC bundle adjustment is to use independent reference GCPs or tie points. The corrected cameras are used to triangulate and reproject these points, and accuracy is reported against reference image coordinates using statistics such as RMSE~\cite{grodecki2003block,fraser2005bias,hanley2004sensor}. This protocol directly measures geolocation accuracy, but requires a set of predefined image observations.

When external ground truth points are unavailable, RPC evaluation can still rely on independent image observations excluded from the adjustment~\cite{ban2024rational,ling2020matching}. Here, a held-out observation denotes one 2D measurement removed from a multi-view tie-point track. After bundle adjustment, the 3D point is triangulated from the remaining observations of the track using the corrected RPCs and reprojected into the image containing the held-out observation, where it is compared with the excluded measurement, as in leave-one-out protocols~\cite{brovelli2008accuracy,irschara2010towards}. Held-out reprojection errors measure cross-view predictive consistency rather than the residual minimized during bundle adjustment. Since the held-out observation is used neither to estimate the camera corrections nor to triangulate the evaluated 3D point, the metric is well suited for cross-method comparison.

In satellite 3D reconstruction pipelines, RPCs can also be evaluated through downstream 3D products, such as digital surface models (DSMs), by comparing reconstructed geometry against LiDAR references~\cite{aati2020optimization,mari2022sat,dangelo2023geometric}. When external elevation data are unavailable, multi-view stereo pipelines rely on consistency-based proxies, such as the dispersion of pairwise height estimates~\cite{dangelo2012dsm,mari2021automatic}. Although they do not measure absolute altitude or geolocation errors, these proxies enable systematic evaluation using only the input imagery and corrected camera models.

Overall, existing protocols form a hierarchy according to the available supervision: independent GCPs provide the strongest evaluation when available, while held-out reprojection and multi-view 3D consistency errors offer practical alternatives for comparing RPC correction methods under GCP-free conditions.
\section{Method}
\label{sec:method}

The proposed bundle adjustment pipeline for RPC refinement of multi-date satellite image collections can be summarized as an 8-step methodology:
\begin{enumerate}
    \item \textit{Feature detection}. A learned local feature extractor is used to detect distinctive keypoints and extract their associated descriptors in each image.
    \item \textit{Image pair selection}. Learned global image descriptors are used to select a reduced set of image pairs for feature matching.
    \item \textit{Pairwise feature matching}. Selected image pairs are matched using a learned sparse matcher robust to seasonal appearance variations.
    \item \textit{Feature track construction}. Pairwise feature correspondences are extended to tracks of length $T$ containing $T \ge 2$ feature observations.
    \item \textit{First bundle adjustment run}. Initial $N_{\text{init}}$ iterations using a soft L1 reprojection loss for robustness to outliers.
    \item \textit{Outlier observation filtering}. Observations with reprojection error above an adaptive threshold are rejected.
    \item \textit{Second bundle adjustment run}. Final $N_{\text{final}}$ outlier-free optimization iterations using an L2 reprojection loss.
    \item \textit{Corrected RPC fitting}. For each input image, the corrected output RPC is fitted from the original RPC and the optimized correction parameters.
\end{enumerate}

As in standard open-source pipelines such as ASP~\cite{beyer2018ames} and SAT-BA{\footnotesize v1}~\cite{mari2021generic}, we use pairwise feature matching to establish point correspondences across the input images (steps 1-4). This is followed by two bundle adjustment runs with an intermediate outlier-observation rejection step. Each bundle adjustment run jointly optimizes the RPC correction parameters and the 3D point coordinates by minimizing reprojection errors with respect to the feature observations. We set the number of optimization iterations to $N_{\text{init}}=50$ and $N_{\text{final}}=300$, and adopt the algorithms of~\cite{mari2021generic} to extend pairwise correspondences into feature tracks and perform adaptive outlier filtering. The RPC correction of each input camera is parameterized by a corrective object-space rotation $R$, defined as in \eqref{eq:sat-ba_formula}, and composed with the original RPC projection function.

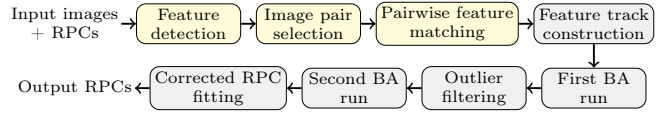
\begin{figure}[t]
\centering
\resizebox{0.7\columnwidth}{!}{%
\begin{tikzpicture}[
    block/.style={
        rectangle,
        draw,
        rounded corners,
        align=center,
        minimum width=1.55cm,
        minimum height=0.55cm,
        inner sep=2pt,
        font=\scriptsize,
        fill=gray!12
    },
    matchblock/.style={
        block,
        fill=yellow!18
    },
    textblock/.style={
        align=center,
        font=\scriptsize,
        inner sep=1pt
    },
    arrow/.style={->, thick},
    node distance=0.35cm and 0.22cm
]

\node[textblock] (in) {Input images\\+ RPCs};
\node[matchblock, right=of in] (b1) {Feature\\detection};
\node[matchblock, right=of b1] (b2) {Image pair\\selection};
\node[matchblock, right=of b2] (b3) {Pairwise feature\\matching};
\node[block, right=of b3] (b4) {Feature track\\construction};

\node[block, below=of b4] (b5) {First BA\\run};
\node[block, left=of b5] (b6) {Outlier\\filtering};
\node[block, left=of b6] (b7) {Second BA\\run};
\node[block, left=of b7] (b8) {Corrected RPC\\fitting};
\node[textblock, left=of b8] (out) {Output RPCs};

\draw[arrow] (in) -- (b1);
\draw[arrow] (b1) -- (b2);
\draw[arrow] (b2) -- (b3);
\draw[arrow] (b3) -- (b4);
\draw[arrow] (b4) -- (b5);
\draw[arrow] (b5) -- (b6);
\draw[arrow] (b6) -- (b7);
\draw[arrow] (b7) -- (b8);
\draw[arrow] (b8) -- (out);

\end{tikzpicture}%
}
\caption{Proposed RPC correction pipeline for multi-date satellite images. Our contributions concern yellow steps: feature detection, image pair selection and feature matching.}
\label{fig:pipeline_overview}
\end{figure}

Fig.~\ref{fig:pipeline_overview} summarizes the complete pipeline. This work focuses on feature extraction, image pair selection, and pairwise matching for automatic tie-point extraction, especially in challenging cross-seasonal cases.

\subsection{Season-Invariant Feature Matching}
\label{sec:matching}

Matching multi-date satellite images with strong appearance changes is challenging. Seasonal variability, snow, thin clouds, shadows or land-cover changes can prevent the reliable extraction of image correspondences~\cite{masquil2026diachronic}. Since the quality of bundle adjustment strongly depends on the availability of consistent correspondences, the matching stage must be robust to these diachronic appearance variations while remaining efficient enough to scale to large image collections.

To select an appropriate matcher, we benchmark several detector-based strategies, i.e., methods that first detect sparse local keypoints and extract descriptors before matching~\cite{xu2024local}, on a set of 155 diachronic pairs. Image pairs were manually selected from DFC2019 \textit{Track3} \mbox{WorldView-3} image crops of $2048\times2048$ pixels covering different areas of Omaha (USA)~\cite{bosch2019semantic, lesaux20192019}. We focus on detector-based methods because they naturally decouple feature extraction from pairwise matching: local features can be computed once per image and reused across multiple image pairs. This is particularly relevant for large-scale bundle adjustment, where each image may be matched against multiple other images. We compare the classical SIFT+FLANN strategy~\cite{lowe2004distinctive, muja2014scalable} with modern learned alternatives, including \mbox{XFeat}~\cite{potje2024xfeat}, \mbox{DeDoDe}~\cite{edstedt2024dedode}, \mbox{ALIKED}~\cite{zhao2023aliked}, \mbox{SuperPoint} \cite{detone2018superpoint}, and SIFT features matched with LightGlue~\cite{lindenberger2023lightglue}. In contrast, detector-free transformer matchers estimate point correspondences directly from image pairs, without precomputed keypoints~\cite{sun2021loftr, wang2024efficient, tuzcuouglu2024xoftr}, but their computational cost on high-resolution crops often necessitates resizing or tiling. Detector-based methods therefore better fit our target setting, where reusable features and scalability are central.

In our benchmark, the number of detected keypoints is limited to the top 10,000 responses according to the detector confidence, and the LightGlue matching confidence threshold is set to $0.2$. All other parameters are kept at their default values. For each method, correspondences are estimated and geometrically verified using RANSAC with a reprojection threshold of $0.3$ pixels. The number of raw matches, verified inliers, inlier ratio, image pairs per second, and peak GPU memory are used as evaluation metrics.\footnote{All experiments reported in this work were run on a single NVIDIA GeForce RTX 3090 GPU with 24 GB of memory.}

\newlength{\matchergap}
\setlength{\matchergap}{1mm}

\newlength{\matcherwidth}
\setlength{\matcherwidth}{\dimexpr(\textwidth-2\matchergap)/3\relax}

\newcommand{\matcherpanel}[2]{%
    \begin{tikzpicture}[inner sep=0, outer sep=0]
        \node[anchor=south west, inner sep=0, outer sep=0] (img)
            {\includegraphics[width=\matcherwidth]{#1}};
        \node[
            anchor=north west,
            fill=white,
            inner sep=1.8pt,
            font=\scriptsize
        ] at ([xshift=2pt,yshift=-2pt]img.north west) {#2};
    \end{tikzpicture}%
}

\begin{figure*}[t]
    \centering
    \setlength{\tabcolsep}{0pt}
    \renewcommand{\arraystretch}{0}

    \begin{tabular}{@{}c@{\hspace{\matchergap}}c@{\hspace{\matchergap}}c@{}}
        \matcherpanel{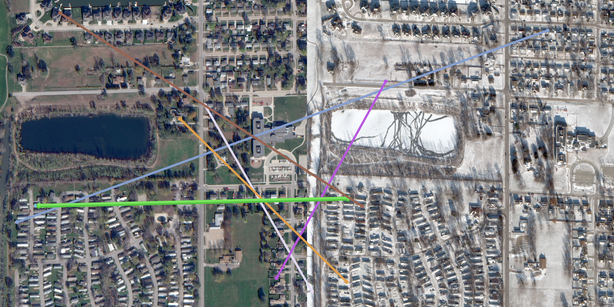}{SIFT+FLANN (9)} &
        \matcherpanel{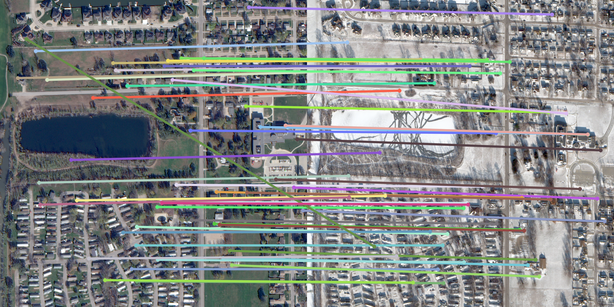}{XFeat (49)} &
        \matcherpanel{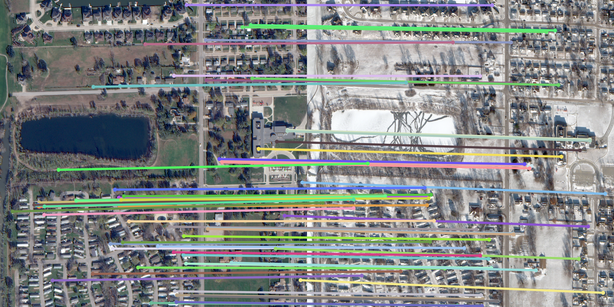}{SIFT+LG (63)}
        \\[\matchergap]
        \matcherpanel{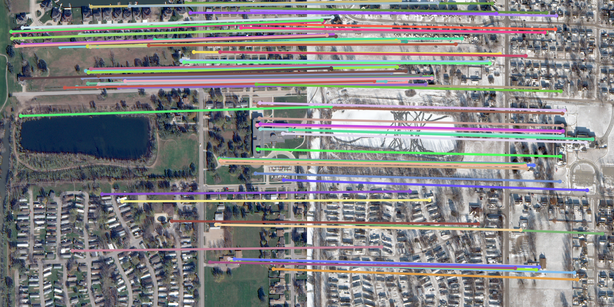}{DeDoDe-B+LG (79)} &
        \matcherpanel{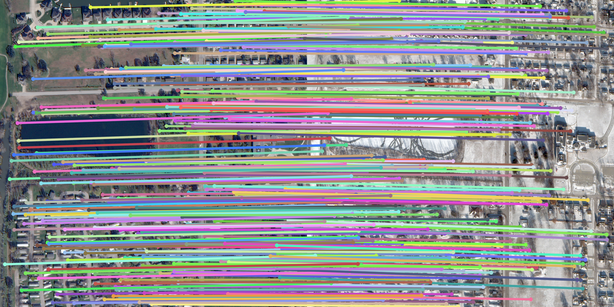}{ALIKED+LG (296)} &
        \matcherpanel{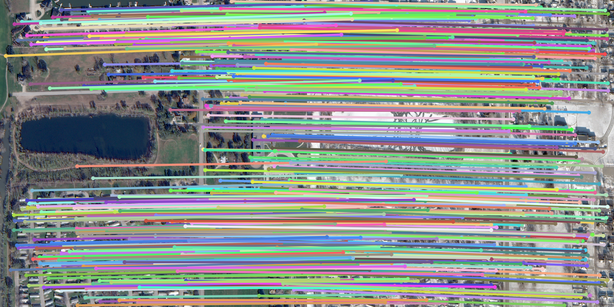}{SP+LG (484)}
    \end{tabular}

    \caption{
        Qualitative comparison of detector-based matchers on a diachronic pair of area \textit{OMA 134}. Parentheses indicate geometrically verified matches (inliers) for this pair; Table~\ref{tab:matching_benchmark} reports averages over 155 pairs. SP+LG yields the most correspondences.
    }
    \label{fig:matching_comparison}
\end{figure*}

\begin{table}[t]
\centering
\caption{Matching benchmark on 155 diachronic satellite image pairs from Omaha. Mean values are reported. Inliers indicate geometrically verified matches.}
\label{tab:matching_benchmark}

\begin{minipage}[c]{0.25\linewidth}
    \centering
    \includegraphics[height=3cm]{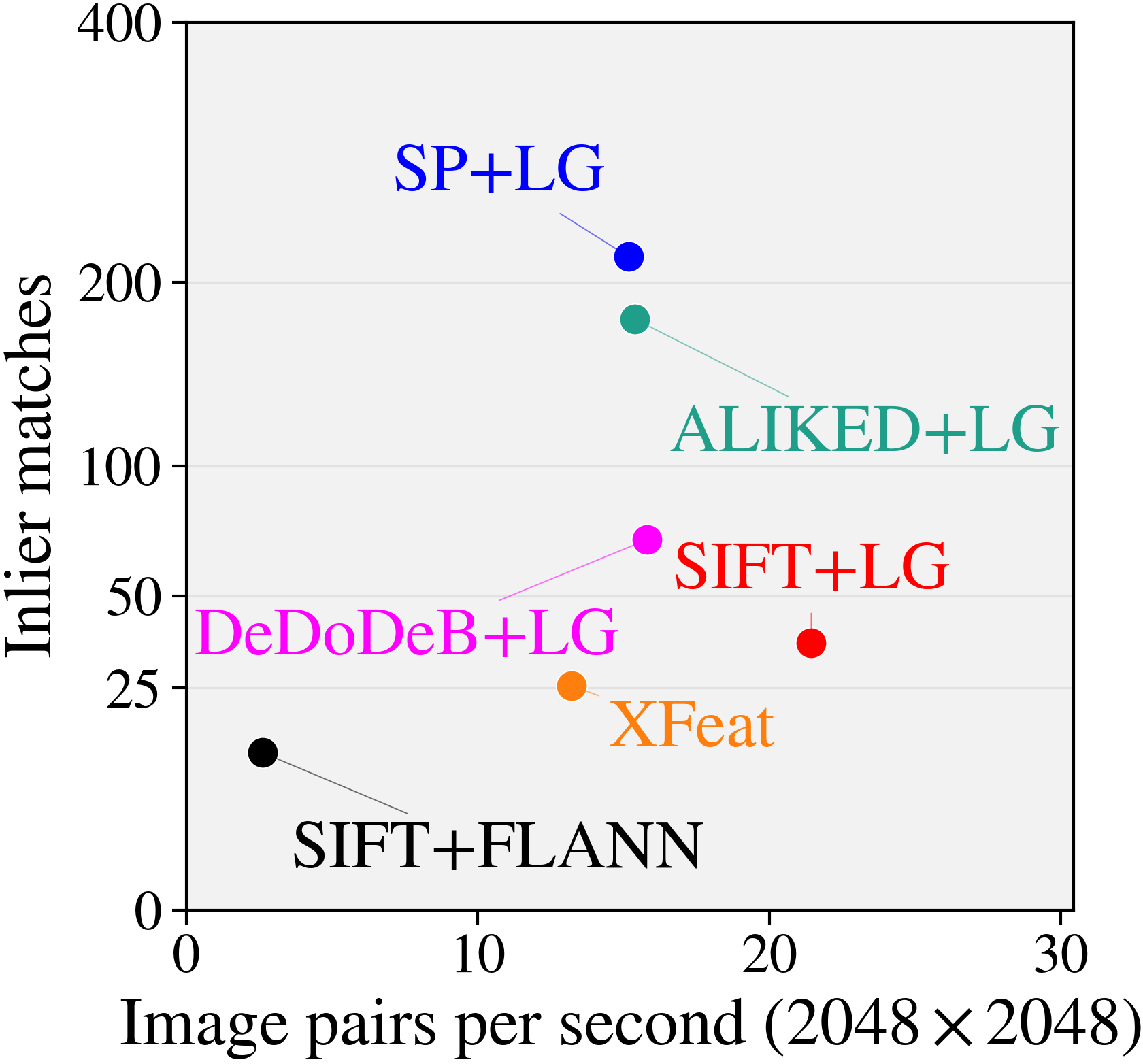}
\end{minipage}
\hfill
\begin{minipage}[c]{0.72\linewidth}
    \centering
    \begin{adjustbox}{width=\linewidth,totalheight=3.0cm,keepaspectratio}
    \begin{tabular}{@{}l@{\hspace{0cm}}ccccc@{}}
    \toprule
    \textbf{Method} & \textbf{Raw matches} & \textbf{Inliers} & \textbf{Inlier ratio} & \textbf{Pairs/s} & \textbf{GPU (MB)} \\
    \midrule
    SIFT+FLANN & 98.39 & 12.57 & 0.128 & 2.62 & \textbf{0.00} \\
    XFeat & \textbf{3163.19} & 25.58 & 0.008 & 13.22 & 783.54 \\
    SIFT+LightGlue & 203.10 & 36.17 & \textbf{0.207} & \textbf{21.44} & 1794.79 \\
    DeDoDe-B+LightGlue & 737.37 & 69.65 & 0.147 & 15.81 & 2849.36 \\
    ALIKED+LightGlue & 2313.27 & 177.23 & 0.099 & 15.39 & 2402.78 \\
    SuperPoint+LightGlue & 2549.88 & \textbf{216.74} & 0.097 & 15.16 & 2451.72 \\
    \bottomrule
    \end{tabular}
    \end{adjustbox}
\end{minipage}
\end{table}

The results in Table~\ref{tab:matching_benchmark} show that SuperPoint+LightGlue provides the best trade-off for efficiently finding season-invariant correspondences. It produces the highest number of geometrically verified correspondences, with 216.74 inliers per pair on average, while maintaining a competitive processing rate of 15.16 image pairs per second. Although SIFT+LightGlue achieves the highest inlier ratio and the fastest runtime, its absolute number of verified correspondences is substantially lower, which limits its usefulness for robust feature-track construction. ALIKED+LightGlue is the closest alternative in terms of final inlier count, but still yields fewer verified matches than SuperPoint+LightGlue. We therefore use SuperPoint+LightGlue as the default matching strategy in our bundle-adjustment pipeline. Fig.~\ref{fig:matching_comparison} shows example correspondences from the benchmarked methods on a diachronic pair.

\begin{figure*}[t]
    \centering
\includegraphics[draft=false, width=\textwidth]{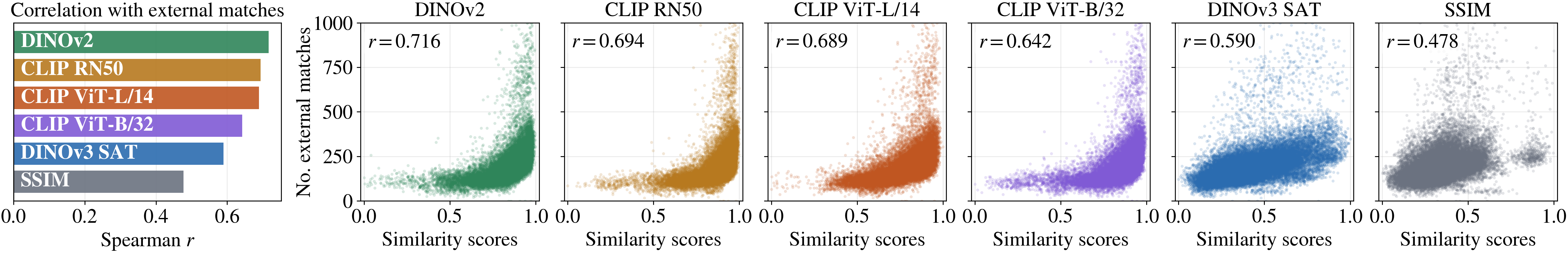}
\caption{
    Correlation between image-pair similarity scores and the number of external matches. Left: Spearman ($r$) correlation for each similarity metric. Right: normalized similarity scores versus external match counts per pair, ordered by decreasing $r$ value.
}
\label{fig:similarity_study}
\end{figure*}

\subsection{Image Pair Similarity Estimation}
\label{sec:image_pair_similarity}

Satellite stereo pair selection has traditionally relied on acquisition metadata,
including viewing angles and temporal proximity~\cite{facciolo2017automatic, qin2019critical}. Recent simulation-based methods improve these heuristics but still rank pairs from metadata-derived viewing geometry~\cite{gomez2023improving}. In multi-date image collections, geometric and temporal metadata are only indirect proxies for visual matchability, since pairs with suitable viewing geometry can still exhibit diachronic appearance changes that degrade local correspondence quality~\cite{masquil2026diachronic}.
We propose a metadata-free criterion for pair selection based on visual content.

Pairwise feature matching is a major bottleneck in large-scale bundle adjustment, and exhaustive matching can introduce noisy correspondences from visually incompatible diachronic pairs~\cite{masquil2026diachronic}. Selecting a reduced subset of visually similar image pairs for feature matching can both reduce runtime and improve the reliability of the resulting feature tracks, provided that the selected pairs still form a connected image graph. To this end, we rank candidate image pairs according to an appearance similarity score.

To identify an appearance similarity score suitable for selecting feature matching pairs in multi-date satellite imagery, we benchmark metrics based on handcrafted and learned global image descriptors on 30 Omaha areas of interest from the DFC2019 \textit{Track3} dataset~\cite{bosch2019semantic, lesaux20192019}, each containing 39-42 diachronic WorldView-3 image crops. We evaluate each metric by comparing its image-pair scores against an independent set of reference pairwise correspondences, obtained for all unordered image pairs using SuperPoint+LightGlue followed by RANSAC filtering at a threshold of $0.3$ pixels.

Six similarity scores are compared, based on SSIM~\cite{wang2004image}, the satellite-pretrained model DINOv3~\cite{simeoni2025dinov3}, the general-purpose model DINOv2~\cite{oquab2024dinov2} and the CLIP~\cite{radford2021learning} variants CLIP RN50, CLIP ViT-B/32, and CLIP ViT-L/14. For SSIM, the pair score is computed using the classic structural similarity index~\cite{wang2004image} clipped to $[0,1]$.
In DINOv2 and DINOv3, each image is represented by the normalized final global-token embedding, while in CLIP variants it is represented by the normalized image embedding returned by the image encoder. For each learned descriptor $f(\cdot)$, the pair score is computed as the cosine similarity
\begin{equation}
c_{ij} =
\frac{f(I_i)^\top f(I_j)}
{\|f(I_i)\|_2 \|f(I_j)\|_2},
\label{eq:cosine_similarity}
\end{equation}
which is mapped from $[-1,1]$ to $[0,1]$ as $s_{ij} = (c_{ij}+1)/2$.

We min-max normalize each similarity metric over the full set of pairs and compute Spearman correlation coefficients~\cite{spearman1904proof}, denoted $r$, between similarity scores and the reference number of pairwise correspondences. These coefficients measure whether higher similarity generally implies more retained matches, without assuming a linear relationship. Results are shown in Fig.~\ref{fig:similarity_study}.

DINOv2-based similarity achieves the strongest alignment with pairwise matchability, reaching a Spearman correlation of $0.716$. It is followed by CLIP RN50, CLIP ViT-L/14, CLIP ViT-B/32, the satellite-pretrained DINOv3 model, and SSIM. The weaker correlation achieved by satellite-pretrained DINOv3 may stem from its stronger invariance to seasonal appearance changes: such invariance can make images with dissimilar local appearance appear close in descriptor space because they remain semantically similar. In contrast, DINOv2 appears to preserve structural and appearance cues that better reflect local feature matchability.

\begin{figure}[t]
    \centering

    \setlength{\tabcolsep}{1.5pt}
    \renewcommand{\arraystretch}{1.0}

    \newcommand{\imagepair}[2]{%
        \hbox to \linewidth{%
            \includegraphics[width=0.5\linewidth]{#1}%
            \includegraphics[width=0.5\linewidth]{#2}%
        }%
    }

    \begin{tabular}{
        >{\centering\arraybackslash}m{0.015\columnwidth}
        >{\centering\arraybackslash}m{0.23\columnwidth}
        >{\centering\arraybackslash}m{0.23\columnwidth}
        >{\centering\arraybackslash}m{0.23\columnwidth}
        >{\centering\arraybackslash}m{0.23\columnwidth}
    }
    
        &
        {\scriptsize\textit{OMA 059}} &
        {\scriptsize\textit{OMA 176}} &
        {\scriptsize\textit{OMA 248}} &
        {\scriptsize\textit{OMA 251}}
        \\[-1pt]
    
        {\rotatebox{90}{\scriptsize Low sim.}} &
        \imagepair
            {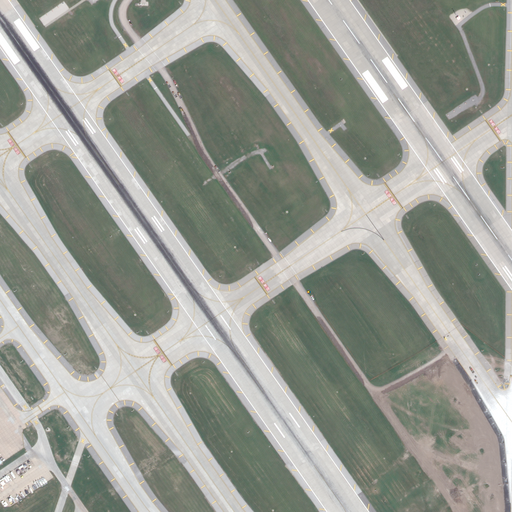}
            {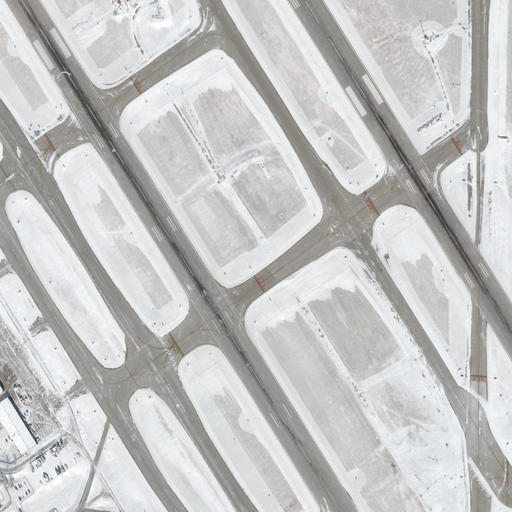}
        &
        \imagepair
            {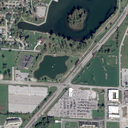}
            {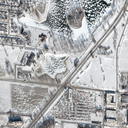}
        &
        \imagepair
            {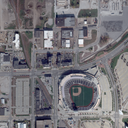}
            {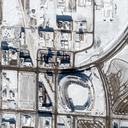}
        &
        \imagepair
            {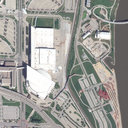}
            {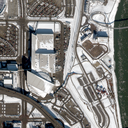}
        \\[1pt]
    
        {\rotatebox{90}{\scriptsize High sim.}} &
        \imagepair
            {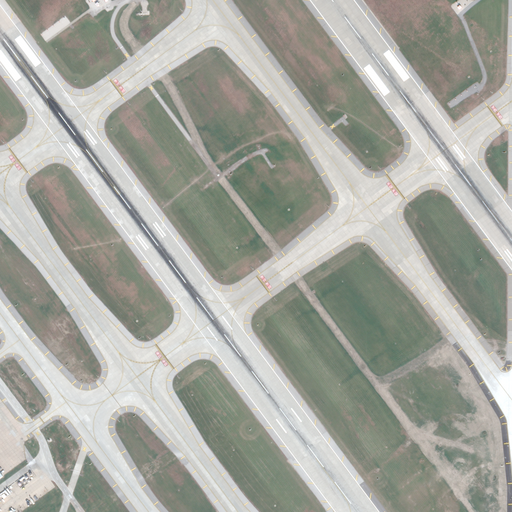}
            {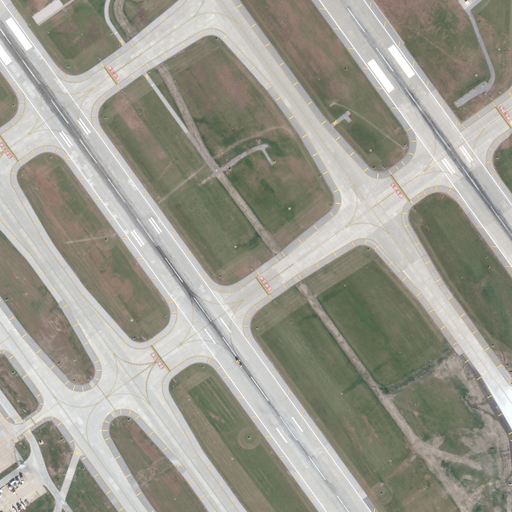}
        &
        \imagepair
            {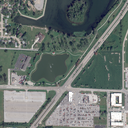}
            {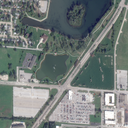}
        &
        \imagepair
            {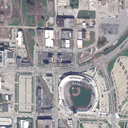}
            {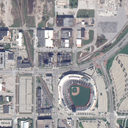}
        &
        \imagepair
            {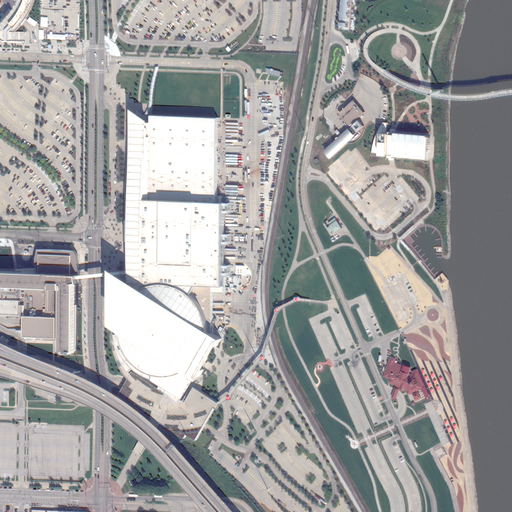}
            {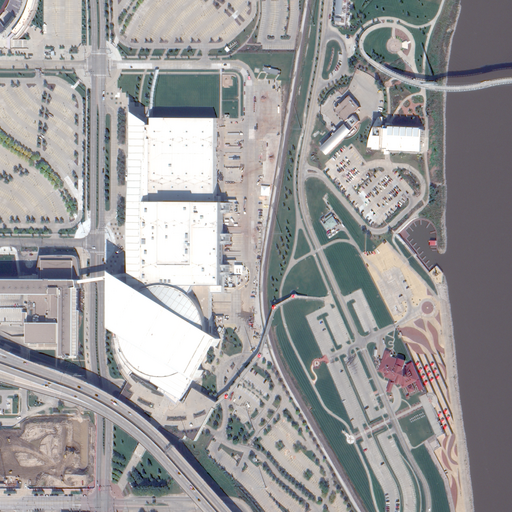}
    
    \end{tabular}

    \caption{
        Qualitative examples of satellite image pairs with high and low DINOv2 similarity scores. The proposed matching strategy prioritizes high similarity pairs.
    }
    \label{fig:dinov2_similarity_examples}
\end{figure}

Based on these findings, we use DINOv2 cosine similarity for selecting suitable image pairs for pairwise feature matching. A low DINOv2 similarity score indicates pairs with stronger seasonal, illumination, or structural appearance changes, whereas a high score indicates pairs expected to yield denser and more reliable local correspondences. Qualitative examples are shown in Fig.~\ref{fig:dinov2_similarity_examples}.

\subsection{Scalable Image Pair Selection via Similarity Scores}
\label{sec:pair_selection}

Given the set of all possible image pairs and their precomputed similarity scores, we formulate pair selection for feature matching as a weighted graph problem. Images are treated as nodes, and candidate pairs as undirected edges weighted by their similarity score. The goal is to retain a few high-similarity edges while preserving connectivity across all images for stable feature track construction and downstream bundle adjustment optimization.

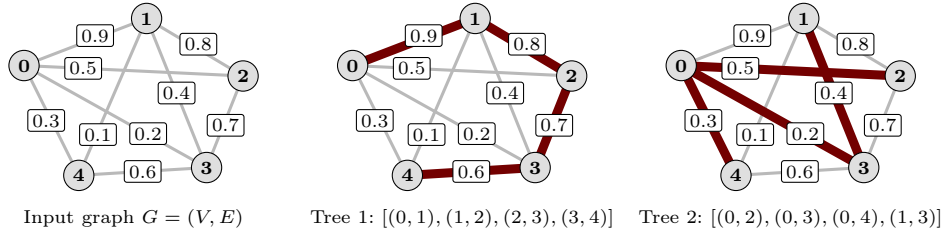
\begin{figure*}[t]
\centering

\begin{tikzpicture}[
    scale=0.69,
    every node/.style={font=\sffamily},
    vertex/.style={
        circle,
        draw=black,
        fill=gray!25,
        minimum size=4mm,
        inner sep=0pt,
        font=\bfseries\scriptsize
    },
    edge/.style={
        draw=gray!55,
        line width=1.15pt
    },
    selected/.style={
        draw=red!45!black,
        line width=3.4pt
    },
    weight/.style={
        draw=black,
        fill=white,
        rounded corners=1pt,
        inner sep=1.8pt,
        font=\scriptsize
    },
    title/.style={
        font=\scriptsize
    },
    footer/.style={
        font=\scriptsize,
        align=center
    }
]

\newcommand{\drawpanel}[4]{%
\begin{scope}[xshift=#1]

    \coordinate (n0) at (0.0, 2.9);
    \coordinate (n1) at (2.35, 3.8);
    \coordinate (n2) at (4.2, 2.7);
    \coordinate (n3) at (3.5, 0.95);
    \coordinate (n4) at (1.05, 0.8);


    \draw[edge] (n0) -- (n1);
    \draw[edge] (n1) -- (n2);
    \draw[edge] (n2) -- (n3);
    \draw[edge] (n3) -- (n4);
    \draw[edge] (n0) -- (n2);
    \draw[edge] (n1) -- (n3);
    \draw[edge] (n0) -- (n4);
    \draw[edge] (n0) -- (n3);
    \draw[edge] (n1) -- (n4);

    #4

    \node[vertex] at (n0) {0};
    \node[vertex] at (n1) {1};
    \node[vertex] at (n2) {2};
    \node[vertex] at (n3) {3};
    \node[vertex] at (n4) {4};

    \node[weight] at ($(n0)!0.58!(n1) + (0.0,0.05)$) {0.9};
    \node[weight] at ($(n1)!0.52!(n2) + (0.03,0.08)$) {0.8};
    \node[weight] at ($(n2)!0.55!(n3) + (0.05,0.02)$) {0.7};
    \node[weight] at ($(n4)!0.50!(n3) + (0.00,0.00)$) {0.6};
    \node[weight] at ($(n0)!0.27!(n2) + (0.0,0.0)$) {0.5};
    \node[weight] at ($(n1)!0.50!(n3) + (0.0,0.0)$) {0.4};
    \node[weight] at ($(n0)!0.50!(n4) + (-0.1, 0.0)$) {0.3};
    \node[weight] at ($(n0)!0.76!(n3) + (-0.28,0.18)$) {0.2};
    \node[weight] at ($(n1)!0.64!(n4) + (-0.1,-0.3)$) {0.1};
    
    \node[footer] at (2.1,-0.02) {#3};

\end{scope}
}

\drawpanel{0cm}
{Starting graph $G$}
{\scriptsize Input graph \(G=(V,E)\)}
{}

\drawpanel{6.3cm}
{\small 1 highlighted}
{\scriptsize Tree 1: $[(0,1),(1,2),(2,3),(3,4)]$}
{
    \draw[selected] (n0) -- (n1);
    \draw[selected] (n1) -- (n2);
    \draw[selected] (n2) -- (n3);
    \draw[selected] (n3) -- (n4);
}

\drawpanel{12.6cm}
{\small Tree 2 highlighted}
{\scriptsize Tree 2: $[(0,2),(0,3),(0,4),(1,3)]$}
{
    \draw[selected] (n0) -- (n2);
    \draw[selected] (n0) -- (n3);
    \draw[selected] (n0) -- (n4);
    \draw[selected] (n1) -- (n3);
}

\end{tikzpicture}

\caption{
Toy example of selecting two edge-disjoint spanning trees ($K=2$) from a matching graph $G=(V,E)$.
Nodes $V$ are input images, edges $E$ denote candidate image pairs for feature matching, and edge weights represent image-pair similarity scores.
The first tree selects the highest-similarity pairs, while the second is computed after removing the edges of the first tree. Edge $(1,4)$ is unused after $K=2$.
}
\label{fig:pair_selection_toy_example}
\end{figure*}

Let \(G=(V,E)\) be the candidate pair graph, where \(V\) is the set of images and \(E\) is the set of edges or candidate pairs. Each edge \((i,j)\in E\) has a similarity score \(s_{ij}\), with larger values indicating more visually compatible image pairs. The selection algorithm iterates to greedily select up to \(K\) edge-disjoint maximum spanning trees. At each iteration, the maximum spanning tree of the current graph is selected, adding its edges to the selected pair set and removing them from the graph. The procedure terminates once \(K\) trees have been selected or the remaining graph is no longer connected. A toy example is shown in Fig.~\ref{fig:pair_selection_toy_example}.

This strategy has two useful properties. First, each selected tree spans all images, so every image remains connected to the rest whenever the candidate graph permits it. Second, extracting multiple spanning trees introduces controlled redundancy: each additional tree contributes an independent set of connections, reducing sensitivity to occasional matching failures while keeping the total number of selected pairs bounded. For \(N\) images, the method selects at most \(K(N-1)\) pairs, yielding linear growth in the number of images rather than the quadratic growth of exhaustive pair matching.

We set \(K=5\) by default, limiting matching to at most five spanning trees of image pairs connecting the input images. This provides a scalable alternative to exhaustive pair matching while preserving graph connectivity, and follows common principles used in scalable SfM systems~\cite{schoenberger2016sfm,lou2012matchminer}, where exhaustive matching is often replaced by an informative subset of image pairs for large collections.

\section{Experiments}
\label{sec:experiments}

\subsection{Data}

The public DFC2019 \textit{Track3} multi-view dataset~\cite{bosch2019semantic, lesaux20192019} is used. Our main benchmark consists of 30 Omaha areas of interest (AOIs), each containing between 39 and 42 diachronic WorldView-3 RGB image crops of size \mbox{$2048 \times 2048$} pixels. Omaha is selected for its pronounced seasonal variability: winter images often contain snow cover, whereas summer acquisitions exhibit markedly different illumination and vegetation conditions. These diachronic changes make the dataset particularly suitable for evaluating the robustness of RPC bundle adjustment under challenging multi-date imagery.

We additionally use 11 Jacksonville AOIs from the same dataset, each containing 24 synchronic WorldView-3 RGB image crops. This subset is used only for a complementary ablation study, where we evaluate whether the proposed strategy preserves performance in the classical synchronic setting.

\subsection{Baseline Methods}

We compare our RPC correction pipeline against two open-source baseline methods: the NASA \textit{Ames Stereo Pipeline} (ASP)~\cite{beyer2018ames} and \textit{sat-bundleadjust} v1 (SAT-BA{\footnotesize v1})~\cite{mari2021generic}. Both baselines implement RPC bundle adjustment and export corrected camera models in RPC format, as required  by the evaluation protocol. For a fair comparison, all methods are evaluated on the same Omaha AOIs, input images and raw RPC models. Both ASP and SAT-BA{\footnotesize v1} use handcrafted feature matching because they do not support learned matching methods.

\subsection{Evaluation Metrics}

We evaluate the corrected RPCs using external pseudo-ground-truth feature tracks that are excluded from all bundle adjustment optimizations. These tracks are automatically constructed using ALIKED+LightGlue, followed by RANSAC-based geometric filtering with a fundamental matrix model. Using a different feature extractor reduces the risk of evaluating the corrected RPCs on feature tracks with the same detector-specific biases used during bundle adjustment. In addition, to make the evaluation sensitive to challenging diachronic appearance changes, we construct the external tracks from visually dissimilar image pairs rather than from the high-similarity pairs selected for bundle adjustment. Specifically, we apply the pair-selection logic of Section~\ref{sec:pair_selection} with reversed edge weights, selecting the $K=5$ spanning trees with minimum DINOv2 similarity scores in each Omaha AOI.

\textbf{Held-out reprojection errors}. For each pseudo-GT feature track, one observation is held out, the 3D point is triangulated from the remaining observations, and then is projected back into the held-out image using its corrected RPC. Errors correspond to Euclidean distances (in pixels) between the held-out observations and the reprojected points. We report mean, median and 90th percentile errors; lower values indicate better cross-view predictive consistency.

\textbf{Pairwise 3D consistency errors}. For each pseudo-GT feature track, every valid pair of observations is used to triangulate a 3D point with the corrected RPCs. Consistent RPCs should produce similar 3D points for the same track. We convert the triangulated points to ECEF coordinates and compute their Euclidean distances, in meters, to a robust center, defined by the component-wise median 3D coordinates of all triangulated points. We report the median and 90th percentile of these consistency distances, as well as the height median absolute deviation (MAD), computed as the median absolute deviation of the pairwise triangulated altitudes within each track. For all metrics, lower values indicate better 3D consistency. The median Euclidean distance measures overall 3D consistency, while the height MAD measures only the vertical consistency.

\begin{table}[t]
\centering
\setlength{\tabcolsep}{4pt}
\caption{Internal stats of RPC correction pipelines (Omaha AOIs). Fastest in red.}
\label{tab:ba_internal_metrics}
\resizebox{0.8\columnwidth}{!}{%
\begin{tabular}{lcccccc}
\toprule
\textbf{Method} & \textbf{Detector+Matcher} & \textbf{\#Pairs} & \textbf{\#Obs/Cam} & \textbf{Err. [px]} & \textbf{Total time [min:s]} \\
\midrule
ASP & SIFT+FLANN & 807.233 & 289.950 & 1.458 & 18:06 \\
SAT-BA{\footnotesize v1} & SIFT+FLANN & 807.233 & 904.874 & 2.353 & 05:45 \\
\rowcolor{red!20}
Ours & SuperPoint+LightGlue & 198.333 & 1670.072 & 0.422 & 02:22 \\
\bottomrule
\end{tabular}}
\end{table}

\begin{table}[t]
\centering
\caption{External feature track evaluation errors (Omaha AOIs). Top accuracy in red.}
\label{tab:heldout_and_3d}
\resizebox{0.8\columnwidth}{!}{%
\begin{tabular}{l@{}p{1em}@{}ccc@{}p{1em}@{}ccc}
\toprule
\multirow{2}{*}{\bf{Method}}
&
&
\multicolumn{3}{c}{\bf{Held-out reprojection errors}}
&
&
\multicolumn{3}{c}{\bf{Pairwise 3D consistency errors}}
\\
\cmidrule(lr){3-5}
\cmidrule(lr){7-9}
&
&
\bf{Mean [px]}
& \bf{Median [px]}
& \bf{P90 [px]}
&
&
\bf{Height MAD [m]}
& \bf{Median [m]}
& \bf{P90 [m]}
\\
\midrule
Raw RPCs & & 1.940  & 1.755 & 3.438 & & 0.590 & 0.781 & 1.786 \\
ASP      & & 1.537  & 1.280 & 2.924 & & 0.474 & 0.625 & 1.443 \\
SAT-BA{\footnotesize v1}   & & 11.086 & 7.141 & 27.505 & & 2.529 & 3.946 & 8.866 \\
\rowcolor{red!20}
Ours
&
& 1.345 & 1.111 & 2.625
&
& 0.399 & 0.555 & 1.234 \\
\bottomrule
\end{tabular}
}
\label{tab:external_metrics}
\end{table}

\subsection{Results}

Table~\ref{tab:ba_internal_metrics} reports internal stats for each RPC correction pipeline, including matching configuration (\textit{Detector+Matcher}), number of matched image pairs, feature observations per camera (\textit{\#Obs/Cam}), internal average reprojection error (\textit{Err.}), and average runtime per AOI. By combining learned feature matching (SuperPoint+LightGlue) with similarity-based pair selection, our pipeline achieves the lowest runtime (02:22), the largest number of observations per camera, and the lowest overall reprojection error on the internally optimized feature tracks.

External pseudo-GT feature tracks further assess geometric accuracy through held-out reprojection and pairwise 3D consistency errors (Table~\ref{tab:external_metrics}). SAT-BA{\footnotesize v1} is ineffective on the input diachronic image sets, yielding higher errors than the original raw RPCs, whereas ASP improves over them. Since both baselines use SIFT+FLANN matching, we attribute this difference to ASP's use of a robust Cauchy loss, which downweights errors from outlier mismatches; replacing it with an L2 loss caused failure on most AOIs.

\begin{figure}[t]
    \centering
    \includegraphics[width=\linewidth]{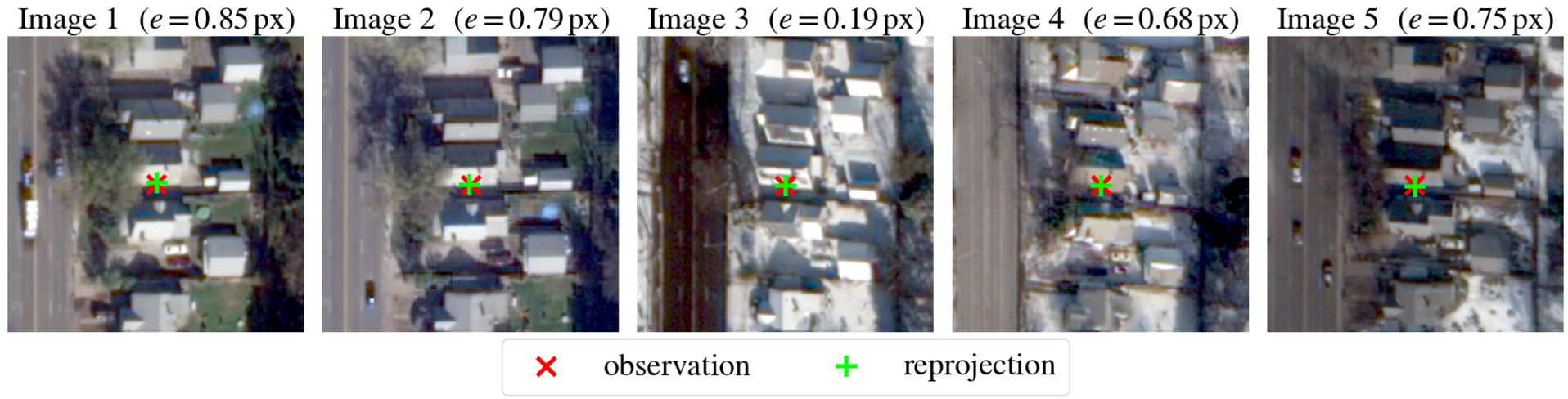}
    \caption{Pseudo-GT feature track from dissimilar diachronic pairs. For evaluation, image 5 is excluded from triangulation, yielding a held-out reprojection error of $0.75$ pixels.}
    \label{fig:diachronic_track}
\end{figure}

Table~\ref{tab:external_metrics} shows that the proposed pipeline achieves the lowest errors across external feature tracks, indicating the most geometrically consistent corrected RPCs. However, absolute magnitudes should be interpreted cautiously, as the pseudo-GT tracks are built from visually dissimilar diachronic pairs and may suffer from localization ambiguity~\cite{kim2024learning}. In addition, held-out reprojection is stricter than the internal BA residuals, since the held-out observation is excluded from 3D point estimation. Together, these factors can explain the larger held-out reprojection errors relative to internal bundle adjustment residuals. Fig.~\ref{fig:diachronic_track} illustrates the held-out error on one pseudo-GT track. Since all methods are evaluated on the same external tracks, the metric remains meaningful for comparing cross-view geometric consistency.

\newcommand{\smalllt}{\raisebox{0.2ex}{${\scriptstyle <}$}}
\newcommand{\smallgt}{\raisebox{0.2ex}{${\scriptstyle >}$}}

\begin{table*}[t]
\centering
\caption{Ablation study on 30 Omaha diachronic AOIs (39-42 images each). Red marks the fastest, lowest-residual configuration succeeding for all cameras.
}
\label{tab:method_comparison}
\resizebox{\textwidth}{!}{%
\begin{tabular}{lccccccccc}
\toprule
\textbf{BA matching configuration} &
\textbf{Matching} &
\textbf{Total time} &
\textbf{\#Pairs} &
\textbf{\#Obs/Cam} &
\textbf{\#Cams} &
\textbf{\#Cams \!\smalllt 1 \!px} &
\textbf{\#Cams \!\smallgt 1 \!px}&
\textbf{\#Cams \!fail} &
\textbf{Err.~\![px]} \\
\midrule
SIFT+FLANN-allpairs & 02:44 & 05:45 & 807.233 & 904.874 & 40.667 & 35.467 & 5.167 & 0.033 & 2.353 \\
SIFT+FLANN-bestpairs ($K$=5) & 00:42 & 03:32 & 198.333 & 612.455 & 40.667 & 40.100 & 0.400 & 0.167 & 0.318 \\
SIFT+LG-allpairs & 03:47 & 07:33 & 807.233 & 3157.709 & 40.667 & 27.967 & 12.700 & 0.000 & 0.938 \\
SIFT+LG-bestpairs ($K$=5) & 00:57 & 03:51 & 198.333 & 1607.234 & 40.667 & 40.633 & 0.033 & 0.000 & 0.447 \\
SP+LG-allpairs & 04:10 & 07:09 & 807.233 & 3505.784 & 40.667 & 33.333 & 7.333 & 0.000 & 0.883 \\
\rowcolor{red!20}
SP+LG-bestpairs ($K$=5) (Ours) & 01:03 & 02:22 & 198.333 & 1670.072 & 40.667 & 40.667 & 0.000 & 0.000 & 0.422 \\
\bottomrule
\end{tabular}%
}
\end{table*}

\begin{table*}[t]
\centering
\caption{Ablation study on 11 Jacksonville synchronic AOIs (24 images each). Red and orange mark the fastest and lowest-residual configurations, respectively.}
\label{tab:method_comparison_2}
\resizebox{\textwidth}{!}{%
\begin{tabular}{lccccccccc}
\toprule
\textbf{BA matching configuration} &
\textbf{Matching} &
\textbf{Total time} &
\textbf{\#Pairs} &
\textbf{\#Obs/Cam} &
\textbf{\#Cams} &
\textbf{\#Cams \!\smalllt 1 \!px} &
\textbf{\#Cams \!\smallgt 1 \!px}&
\textbf{\#Cams \!fail} &
\textbf{Err.~\![px]} \\
\midrule
SIFT+FLANN-allpairs & 00:58 & 02:37 & 276.000 & 957.205 & 24.000 & 24.000 & 0.000 & 0.000 & 0.327 \\
\rowcolor{orange!20}
SIFT+FLANN-bestpairs ($K$=5) & 00:25 & 02:04 & 115.000 & 782.064 & 24.000 & 24.000 & 0.000 & 0.000 & 0.254 \\
SIFT+LG-allpairs & 01:20 & 03:15 & 276.000 & 2892.693 & 24.000 & 23.909 & 0.091 & 0.000 & 0.699 \\
SIFT+LG-bestpairs ($K$=5) & 00:34 & 02:20 & 115.000 & 1745.451 & 24.000 & 24.000 & 0.000 & 0.000 & 0.420 \\
SP+LG-allpairs & 01:26 & 02:30 & 276.000 & 2491.485 & 24.000 & 24.000 & 0.000 & 0.000 & 0.488 \\
\rowcolor{red!20}
SP+LG-bestpairs ($K$=5) (Ours) & 00:37 & 01:19 & 115.000 & 1557.519 & 24.000 & 24.000 & 0.000 & 0.000 & 0.354 \\
\bottomrule
\end{tabular}%
}
\end{table*}

\subsection{Ablation Study}

We conduct an ablation study to evaluate the impact of the main components of the proposed matching strategy. Table~\ref{tab:method_comparison} reports average results across the 30 Omaha AOIs for different detector+matcher combinations and pair selection strategies. It includes pairwise matching and full-pipeline runtimes, the average number of matched image pairs (\textit{\#Pairs}), feature observations per camera (\textit{\#Obs/Cam}), input cameras (\textit{\#Cams}), cameras with final average reprojection error below or above one pixel, cameras disconnected from the matching graph (\textit{\#Cams~fail}), and the final internal average reprojection error or residual (\textit{Err.}).

Table~\ref{tab:method_comparison} shows that exhaustive pair matching is both slower and less robust than the proposed similarity-based pair selection strategy. Across all configurations, pair selection with $K=5$ consistently reduces runtime and improves the final reprojection error, as shown by the \textit{-allpairs} and \textit{-bestpairs} variants. This result indicates that restricting matching to visually compatible image pairs improves scalability while avoiding noisy correspondences from dissimilar pairs.

\textit{SP+LG-bestpairs} achieves the best trade-off among evaluated configurations: it obtains the lowest total runtime (02:22), keeping all cameras below one pixel reprojection error without any disconnected cameras. Although \textit{SIFT+FLANN-bestpairs} obtains the lowest average reprojection error, it produces fewer observations per camera and occasionally disconnects cameras (\textit{\#Cams~fail} $>$ 0), yielding a less stable matching graph. Overall, the ablation confirms that both robust learned feature matching and similarity-based pair selection are important for efficient and reliable RPC bundle adjustment in diachronic satellite imagery.

We further evaluate our approach in a classical synchronic setting using the 11 Jacksonville AOIs. Table~\ref{tab:method_comparison_2} reports average results across AOIs. Exhaustive SIFT+FLANN matching already performs well in this scenario, yielding reliable connectivity and subpixel reprojection errors. Nevertheless, the proposed pair selection strategy reduces matching time without degrading bundle adjustment performance. \textit{SIFT+FLANN-bestpairs} achieves lower final reprojection residual than \textit{SP+LG-bestpairs} ($0.254$ versus $0.354$ pixels), which may reflect the higher subpixel localization accuracy of SIFT keypoints under ideal conditions~\cite{kim2024learning}.

\section{Conclusion}
\label{sec:conclusion}

We presented a robust RPC bundle adjustment pipeline for multi-date satellite imagery with strong seasonal appearance variation. The proposed method improves automatic tie-point extraction by combining SuperPoint features with LightGlue matching to establish season-invariant correspondences across the input views. DINOv2-based image-pair similarity scores are further used to select an informative subset of image pairs, reducing redundant and error-prone matching while preserving graph connectivity.

Experiments on seasonally diverse WorldView-3 RGB imagery show that our approach improves RPC refinement over open-source baselines, achieving lower held-out reprojection and 3D consistency errors with reduced runtime. Feature-matching and image-similarity benchmarks support the use of SuperPoint+LightGlue and DINOv2, while ablations show that the proposed strategy also reduces matching time without degrading performance in synchronic collections. Overall, the proposed method provides an effective and scalable path toward reliable RPC refinement of large multi-date satellite image archives.


%
%
\bibliographystyle{splncs04}
\bibliography{main, our_refs, matchers_refs}
\end{document}